# Evaluation of Medical Vision Language Models HuluMed and MedGemma, and general purpose chatbots Gemma 3, ChatGPT Plus, and Claude Pro on real previously unseen wound images

Yunzhe Xue‡, Mohammed Saim Ahmed Quadri*, Neal Panse†, Justin W. Ady†, and Usman Roshan‡
* Department of Computer Science, New Jersey Institute of Technology, Newark, NJ, USA
mohammedsaimquadri@gmail.com
† Vascular and Endovascular Surgery, Robert Wood Johnson Hospital, New Brunswick, NJ, USA
nsp112@rwjms.rutgers.edu; jwa60@rwjms.rutgers.edu
‡ Department of Data Science, New Jersey Institute of Technology, Newark, NJ, USA
yunzhexue@gmail.com; usmanroshan@gmail.com

*Abstract*—**Chronic wound assessment remains a clinically challenging task that requires accurate interpretation of wound morphology, tissue composition, vascular characteristics, and infection risk. Recent advances in Vision-Language Models (VLMs) have introduced the possibility of automated multimodal wound analysis through image understanding combined with clinical reasoning. This study evaluates the performance of several general-purpose and medically specialized open–source and proprietary VLMs for clinical wound assessment using an expanded, curated dataset of 20 clinically diverse wounds spanning vascular, surgical, ischemic, venous, lymphedema, and amputation-related etiologies. Six VLMs were evaluated using a structured twelve-question clinical framework covering wound classification, infection risk, vascular intervention recommendations, debridement urgency, wound therapy selection, and advanced management planning. Across 20 wound cases and 240 clinician-graded wound-analysis decisions, ChatGPT achieved the highest overall performance with 174/240 correct responses (72.50%), followed by Claude with 149/240 (62.08%). Among the open–source and medically specialized models, HuluMed achieved the strongest performance with 96/240 correct responses (40.00%), followed by Gemma 3 (81/240, 33.75%), MedGemma 4B (62/240, 25.83%), and MedGemma 27B (42/240, 17.50%). The findings suggest that frontier general-purpose multimodal systems currently demonstrate substantially stronger wound-analysis performance than medically specialized alternatives, highlighting the continued importance of broad multimodal reasoning capabilities alongside domain-specific medical knowledge. Although current VLMs demonstrate promising potential for clinical decision support, substantial limitations remain in advanced wound-management reasoning, procedural planning, and autonomous clinical reliability.**

## I. Introduction

Chronic wounds represent a major healthcare burden worldwide, affecting millions of patients annually and contributing substantially to morbidity, hospitalization, and healthcare expenditure [16], [17]. Accurate wound assessment is critical for treatment planning, vascular intervention decisions, infection management, and monitoring healing progression. However, wound evaluation often depends heavily on clinician expertise and subjective interpretation, leading to variability in management decisions.

Artificial Intelligence (AI) has increasingly been explored as a tool for assisting clinicians in medical image interpretation. Earlier approaches primarily relied on convolutional neural networks (CNNs) for wound segmentation, diabetic foot ulcer detection, and tissue classification tasks [14], [15], [18]. While these systems demonstrated promising performance for narrow prediction tasks, they lacked the broader reasoning capabilities required for clinically meaningful wound-management support.

The emergence of Vision-Language Models (VLMs) has significantly expanded multimodal AI capabilities [19]–[22]. Unlike traditional computer vision systems, VLMs combine image understanding with natural language reasoning, enabling clinically relevant interactions such as infection assessment, intervention recommendations, wound characterization, and therapy planning.

Medical VLMs such as HuluMed [3] and MedGemma [2] were specifically developed to improve performance on healthcare tasks through domain-specific fine-tuning. However, it remains unclear whether medical specialization consistently improves wound-care reasoning compared to broader general-purpose multimodal systems. This study evaluates multiple VLMs using a structured wound-analysis benchmark involving clinically oriented management questions across 20 diverse wound categories.

## II. Related Work

Medical vision-language modeling has evolved rapidly due to advances in transformer architectures, large language models, and multimodal pretraining [19]–[21].

Systems such as LLaVA-Med [1], MedGemma [2], and Hulu-Med [3] have been developed to bridge visual medical interpretation with clinical language reasoning, enabling image understanding, medical visual question answering, and clinical decision-support applications. More recently, open multimodal foundation models such as Gemma 3 [4], [19]–[21] have further expanded the accessibility of large-scale vision-language systems for healthcare research.

Previous AI-assisted wound-analysis systems primarily focused on segmentation, ulcer detection, tissue classification, and wound measurement using conventional computer vision and deep learning techniques [14], [15], [18]. While these approaches demonstrated utility in controlled settings, they generally lacked the ability to perform contextual clinical reasoning, integrate patient-specific considerations, or generate management recommendations. Consequently, their applicability to complex wound-care decision making remained limited.

Several recent studies have attempted to systematically evaluate medical vision-language models across a variety of clinical domains. Benchmark datasets such as VQA-RAD [12], SLAKE [13], PathVQA [11], and PMC-VQA [10] have played a central role in evaluating medical visual question answering systems, while models such as Med-Flamingo [9] demonstrated the potential of large-scale multimodal learning for clinical image interpretation. Liu et al. [7] conducted a comprehensive benchmarking study comparing general-purpose and medically specialized vision-language models across multiple medical visual question-answering and reasoning benchmarks. Their findings demonstrated that large general-purpose multimodal models frequently matched or exceeded the performance of medically specialized systems, while reasoning tasks consistently remained more challenging than image-understanding tasks. Importantly, the authors concluded that no evaluated model achieved reliability sufficient for clinical deployment, highlighting the need for improved multimodal reasoning and more rigorous evaluation methodologies.

Similarly, Kurz et al. [8] evaluated open-source vision-language models and GPT-4o on the NEJM Image Challenge dataset in emergency and critical care settings. Their study reported a substantial performance gap between proprietary and open-source systems, with GPT-4o achieving 68.1% diagnostic accuracy compared with a maximum accuracy of 40.4% among the evaluated open-source models. The authors concluded that current open-source vision-language models require additional domain-specific training and optimization before they can be considered suitable for real-world clinical decision-support applications.

Despite these advances [16], [17], wound care remains comparatively underrepresented in the medical vision-language literature. Most existing benchmarks focus on radiology, pathology, emergency diagnostics, or general medical visual question answering. Furthermore, many evaluations rely on curated benchmark datasets rather than clinically realistic management-oriented assessment frameworks. In contrast, our study evaluates both medically specialized and general-purpose multimodal systems using real previously unseen wound images and a clinically structured twelve-question assessment framework. Beyond diagnosis alone, our evaluation examines wound classification, vascular intervention decisions, infection assessment, debridement planning, dressing selection, and wound-management reasoning, thereby providing a benchmark that more closely reflects real-world wound-care decision support.

## III. Data and Evaluation Framework

### A. Dataset Selection and Distribution

The evaluation dataset consists of 20 clinically diverse wound cases representing multiple wound etiologies and anatomical locations. The wound images were obtained from real patients by the collaborating medical team under Institutional Review Board (IRB) approval and were de-identified prior to inclusion in this study to protect patient privacy. The distribution of wound categories evaluates a model's robustness across highly heterogeneous tissue compositions and complex presentation matrices. The categorical configuration of the baseline cohort is detailed in Table I.

TABLE I
Dataset Distribution by Wound Etiology

| Wound Type / Category | Case Identifiers | Total Cases |
|---|---|---|
| Diabetic / Neuropathic Ulcer | Case 1 | 1 |
| Surgical / Postoperative Wound | Case 2, Case 4 | 2 |
| Ischemic / Arterial Insufficiency | Case 3, Case 11, Case 14, Case 20 | 4 |
| Amputation Stump / Amputation-related Wound | Case 5, Case 8, Case 10, Case 18 | 4 |
| Arteriovenous Fistula Complication | Case 6 | 1 |
| Venous Lower-Extremity Ulcer | Case 7, Case 16, Case 19 | 3 |
| Lower-Extremity Fasciotomy Wound | Case 9, Case 12 | 2 |
| Lymphedema-Associated Wound | Case 13 | 1 |
| Subcutaneous Hematoma | Case 15 | 1 |
| Infected Toe Wound | Case 17 | 1 |
| **Total Benchmark Cohort** | **Cases 1–20** | **20** |

### B. Clinical Evaluation Framework

Each wound image was evaluated using a standardized twelve-question clinical framework designed to assess both direct visual interpretation and higher-order clinical reasoning. The targeted structural checklist parameters are mapped in Table II.

### C. Clinical Reference Standard

To establish a definitive ground truth, a Clinical Reference Standard was developed. Each of the 20 wound cases was independently annotated and resolved by a panel of wound-care experts. The expert answers provide specific textual criteria for each question (e.g., specifying "Betadine" as the exact required dressing for stable dry ischemic gangrene or identifying the precise debridement urgency for infected structural defects). Model outputs were rigorously evaluated against these predefined reference standards, eliminating the subjective ambiguity common in open-vocabulary assessment.

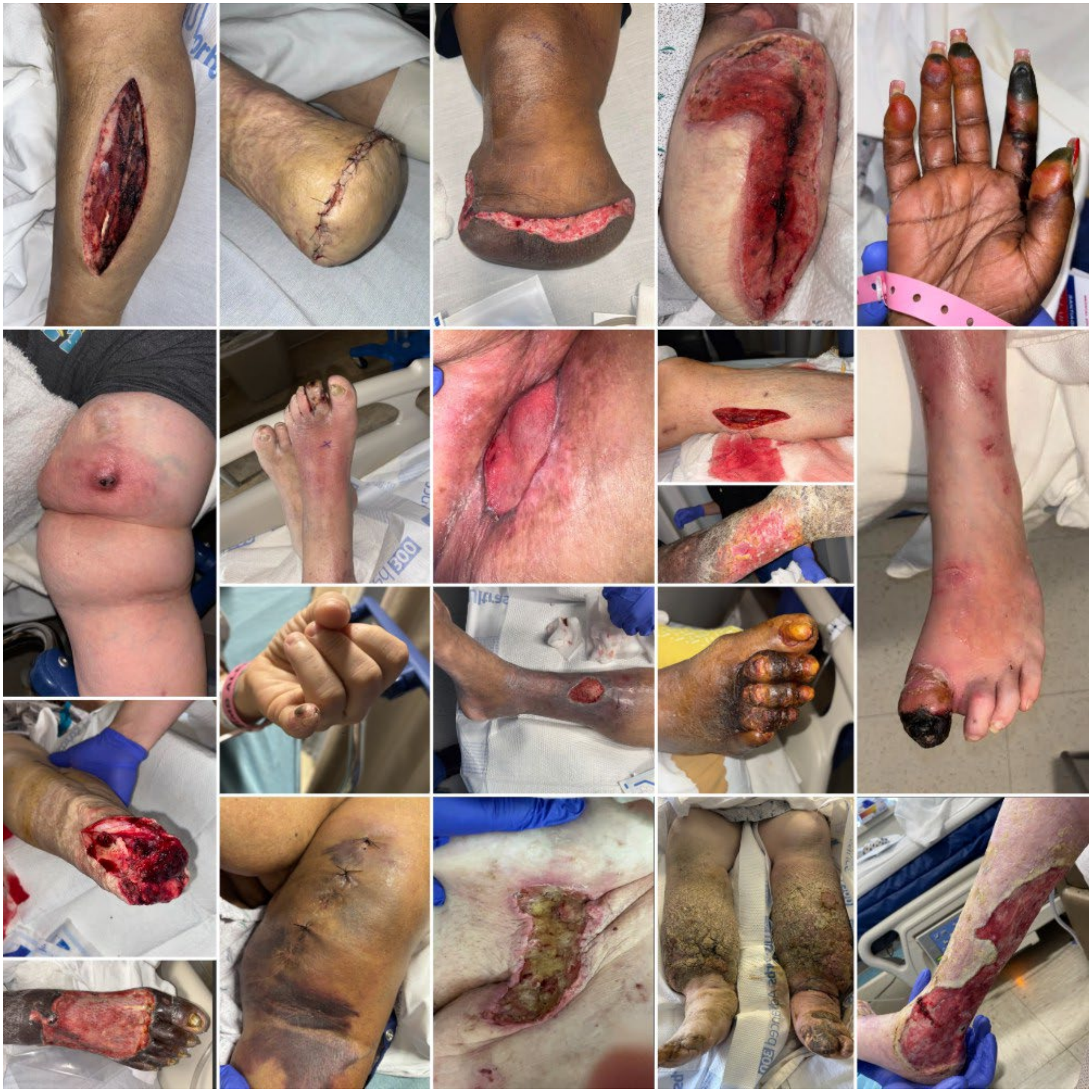

Fig. 1. Representative wound dataset comprising 20 clinically diverse wound presentations spanning diabetic, ischemic, venous, surgical, fasciotomy, lymphedema, fistula-related, and amputation-associated wounds.

### D. Evaluated Models

The evaluation matrix features six core architectures split by deployment configuration:

- **Open–Source / Medically Tailored Models:** HuluMed [3], Gemma 3 [4], MedGemma 27B [2], and MedGemma 4B [2].
- **Proprietary / Frontier Generalist Systems:** Claude (Anthropic) [5] and ChatGPT (OpenAI) [6], accessed

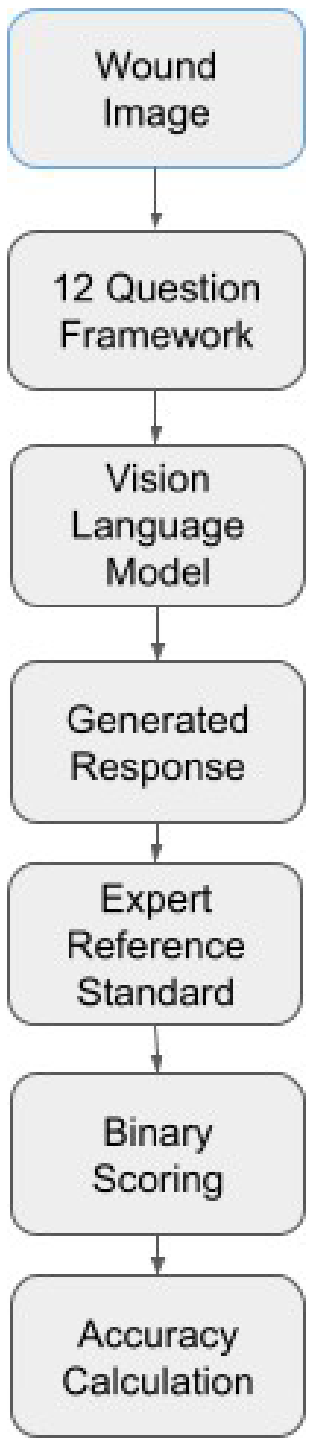

Fig. 2. Overview of the evaluation pipeline.

through their publicly available interfaces during the evaluation period.

All six architectures processed identical image-question prompt sets under zero-shot conditions, generating a validation matrix of 240 total distinct scored evaluation vectors.

### *E. Scoring Methodology*

Each architecture received the raw wound image alongside the standardized 12-question text prompt block. Responses were parsed and evaluated under a clinician-guided binary framework where credit (Score = 1) was awarded exclusively if the model matched the absolute medical criteria defined by the expert reference standard.

Any open-ended response that exhibited hedging, structural ambiguity, or non-committal explanations was automatically coded as incorrect (Score = 0).This evaluation framework therefore measured not only clinical correctness but also a model's ability to provide clear and actionable recommendations. Models that correctly identified the clinical pathway but expressed substantial uncertainty or conditional reasoning were penalized under the binary scoring scheme. For example, if a model suggested "imaging could be considered depending on clinical signs" instead of declaring a definitive modality or "Not at this time," it failed to meet the threshold of actionable decision-making and received no credit.

TABLE II
12-QUESTION CLINICAL MATRIX FRAMEWORK

| ID | Evaluation Domain | Target Assessment Parameter |
|---|---|---|
| Q1 | Etiology Classification | Diagnostic classification (Arterial, Venous, Neither) |
| Q2 | Vascular Assessment | Indication for vascular intervention (Yes/No) |
| Q3 | Morphological Analysis | Presence of excessive fibrinous tissue (Yes/No) |
| Q4 | Morphological Analysis | Healthy granulation tissue coverage >50% (Yes/No) |
| Q5 | Pathology Triage | Potential concern for active infection (Yes/No) |
| Q6 | Diagnostic Planning | Need for axial imaging modality (MRI, CT, etc.) |
| Q7 | Procedural Planning | Debridement setting requirement (OR, Bedside, Neither) |
| Q8 | Prognostic Reasoning | Amputation risk level without urgent intervention |
| Q9 | Advanced Management | Optimal choice of therapeutic dressing |
| Q10 | Advanced Management | Indications for compression therapy (Yes/No) |
| Q11 | Advanced Management | Benefit from negative pressure therapy (VAC) |
| Q12 | Advanced Management | Suitability for bioengineered skin substitute/graft |

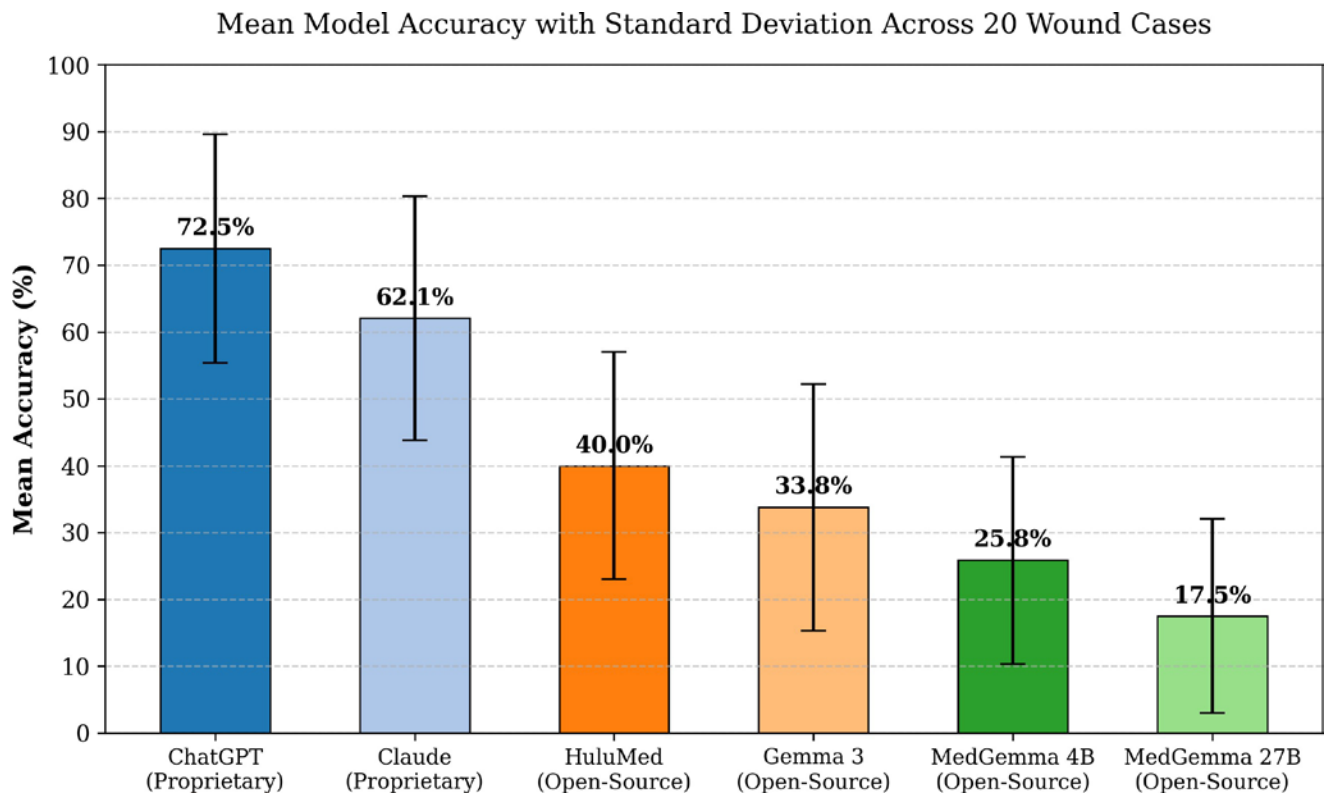

Fig. 3. Mean model accuracy across 20 wound cases with one standard deviation error bars showing performance variability.

## IV. RESULTS

### *A. Overall Performance*

Evaluating the six model configurations across all 20 wound cases yielded a clear performance hierarchy, detailed in Table III. Across 240 clinician-graded wound-assessment decisions, ChatGPT achieved the highest overall performance with 174 correct responses (72.50%), followed by Claude with 149 correct responses (62.08%). Among the open-source systems, HuluMed demonstrated the strongest performance with 96 correct responses (40.00%), followed by Gemma 3 with 81 correct responses (33.75%), MedGemma 4B with 62 correct responses (25.83%), and MedGemma 27B with 42 correct responses (17.50%).

TABLE III
OVERALL MODEL PERFORMANCE ACROSS 240 CLINICAL DECISIONS

| Model | Correct Responses | Accuracy (%) |
|---|---|---|
| ChatGPT | 174/240 | 72.50 |
| Claude | 149/240 | 62.08 |
| HuluMed | 96/240 | 40.00 |
| Gemma 3 | 81/240 | 33.75 |
| MedGemma 4B | 62/240 | 25.83 |
| MedGemma 27B | 42/240 | 17.50 |

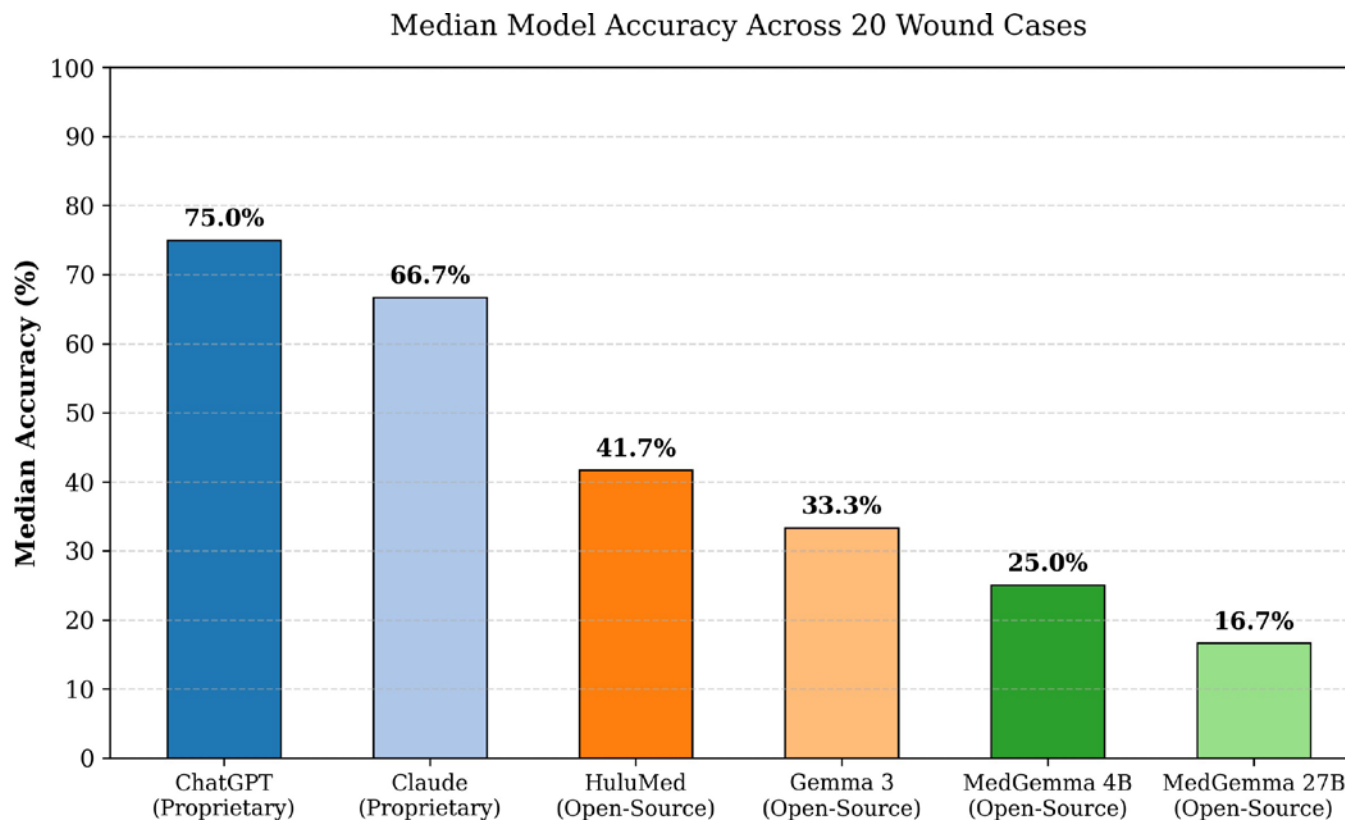

Fig. 4. Median model accuracy across 20 wound cases, illustrating the central tendency of performance for each model.

The proprietary frontier systems substantially outperformed all evaluated open-source alternatives. ChatGPT achieved the highest clinical convergence and demonstrated the strongest consistency across diverse wound presentations. Claude maintained competitive performance but exhibited slightly greater variability across cases. Within the open-source cohort, HuluMed emerged as the strongest performer, while the MedGemma variants showed persistent difficulties translating medical specialization into reliable wound-management reasoning.

### *B. Performance Variability Analysis*

TABLE IV
PER-CASE PERFORMANCE STATISTICS ACROSS 20 WOUND CASES

| Model | Mean Accuracy (%) | Median Accuracy (%) |
|---|---|---|
| ChatGPT | 72.50 | 75.00 |
| Claude | 62.08 | 66.67 |
| HuluMed | 40.00 | 41.67 |
| Gemma 3 | 33.75 | 33.33 |
| MedGemma 4B | 25.83 | 25.00 |
| MedGemma 27B | 17.50 | 16.67 |

To assess performance consistency across wound presentations, we computed per-case mean accuracies, median accuracies, and standard deviations using all twenty wound images. ChatGPT achieved the highest mean accuracy (72.50%) and median accuracy (75.00%), followed by Claude with a mean accuracy of 62.08% and median accuracy of 66.67%. Among the open-source models, HuluMed demonstrated the strongest performance with a mean accuracy of 40.00% and median accuracy of 41.67%.

Standard deviations ranged from 14.53% to 18.43%, as illustrated by the error bars in Figure 3. Although variability was observed across different wound presentations, the proprietary systems consistently outperformed the evaluated open-source alternatives.

### *C. Wound Etiology Analysis*

The benchmark cohort encompassed a diverse range of wound etiologies, including ischemic wounds, venous ulcers, postoperative defects, fasciotomy wounds, amputation-related wounds, lymphedema-associated wounds, vascular access complications, hematomas, and infected wounds. This diversity enabled evaluation of model performance across both common and atypical clinical presentations.

Across wound categories, proprietary models consistently outperformed open-source alternatives. Ischemic wounds with characteristic features such as dry gangrene, tissue necrosis, and distal discoloration were generally recognized more reliably than wounds requiring nuanced contextual interpretation. In contrast, postoperative wounds, fasciotomy defects, and complex healing wounds frequently produced errors related to treatment selection, debridement recommendations, and procedural planning.

Atypical presentations, including lymphedema-associated wounds, hematomas, and partially healed or closed surgical wounds, represented some of the most challenging scenarios. These cases often required interpretation of underlying pathology and longitudinal wound status rather than simple recognition of visible tissue characteristics. Performance degradation in these categories suggests that current vision-language models remain substantially stronger at visual pattern recognition than at comprehensive clinical wound assessment and management planning.

### *D. Question-Level Analysis*

The use of strict binary scoring highlighted a major difference in how these models generate clinical text. The grading framework penalized linguistic ambiguity and required clear clinical choices. Claude's overall accuracy score (62.08%) was heavily impacted by this constraint. A review of its outputs reveals that Claude frequently identified the correct clinical path but paired its choice with protective language, nested conditional modifiers, and extensive clinical qualifiers. While these nuances align with defensive clinical charting, they were scored as incorrect under a framework that requires direct, binary decisions.

Analyzing model responses by question type revealed a consistent distinction between visual interpretation tasks and higher-level clinical decision-making tasks.

- **Visual Assessment Tasks:** Models generally performed better when identifying directly observable wound characteristics, including tissue appearance, necrosis, granulation tissue, fibrinous burden, and visible signs of infection.
- **Clinical Management Tasks:** Performance declined when responses required translating visual findings

TABLE V
EXAMPLE OF CLINICAL AGREEMENT BETWEEN FRONTIER MODELS AND REFERENCE STANDARD

| Assessment Category | Reference Standard | ChatGPT Response Summary | Claude Response Summary | Outcome |
|---|---|---|---|---|
| Q1: Etiology Classification | Arterial | "...dry necrotic gangrene confirms an arterial presentation." | "...tissue death secondary to arterial insufficiency." | Match (1) |
| Q2: Vascular Intervention | Yes, arterial | "Yes, arterial revascularization is strictly indicated." | "Yes, an immediate arterial evaluation is vital." | Match (1) |
| Q5: Active Infection Risk | Yes | "Yes, the surrounding localized erythema indicates risk..." | "Yes, there is definite concern regarding active pathosis..." | Match (1) |
| Q9: Optimal Local Dressing | Betadine | "Apply betadine paint to maintain protective desiccation." | "Dry gangrene should be treated with betadine." | Match (1) |

TABLE VI
EXAMPLE OF CLINICAL FAILURE MODES IN MEDGEMMA MODELS

| Assessment Category | Reference Standard | MedGemma 27B Response | MedGemma 4B Response | Error Profile |
|---|---|---|---|---|
| Q5: Active Infection Concern | No | "Yes, the visual appearance shows severe localized signs..." | "Yes, immediate antibiotic intervention is necessary..." | False Positive Misinterpretation |
| Q6: Axial Imaging Selection | Not at this time | "An emergency MRI should be performed immediately." | "Order a lower extremity CT scan right away." | Over-Testing Recommendation |
| Q7: Debridement Selection | Neither | "Perform sharp aggressive debridement at the bedside." | "Schedule patient for urgent surgical debridement in the OR." | Critical Clinical Failure |

into management decisions, such as determining the need for vascular evaluation, selecting appropriate imaging modalities, recommending debridement strategies, identifying suitable dressings, or assessing the appropriateness of compression therapy, wound vacuum therapy, and skin-substitute application.

These findings suggest that current vision-language models possess stronger capabilities in wound feature recognition than in the synthesis of those findings into comprehensive clinical management recommendations. Notably, the largest performance differences between models emerged in management-oriented questions involving vascular intervention, imaging selection, debridement planning, compression therapy, negative-pressure wound therapy, and skin-substitute recommendations. In contrast, most models performed comparatively better on direct visual interpretation tasks such as tissue characterization, granulation assessment, and infection recognition.

HuluMed demonstrated the most consistent performance among the open–source models, frequently identifying major wound characteristics correctly but often struggling with procedural planning, imaging recommendations, and debridement selection. Gemma 3 exhibited stronger general reasoning than the MedGemma variants but lacked the wound-specific consistency demonstrated by HuluMed. The MedGemma models performed poorly on this benchmark, frequently generating verbose responses, contradictory recommendations, etiological misclassification, and inappropriate management plans. MedGemma 27B in particular struggled with structured prompts and often failed to produce direct, actionable answers, contributing to its low overall accuracy of 17.50%.

### *E. Expert Qualitative Assessment*

Beyond quantitative scoring, model outputs were reviewed qualitatively to assess clinical coherence, management reasoning, and alignment with expert wound-care practice. Qualitative review was informed by expert evaluation and comparison of model-generated responses against the clinical reference standard.

ChatGPT and Claude consistently produced the most clinically coherent responses. These models demonstrated stronger integration of wound morphology, vascular assessment, infection risk evaluation, and management planning. Their recommendations were generally more aligned with expert expectations across diverse wound etiologies.

TABLE VII
COMMON CLINICAL FAILURE CATEGORIES OBSERVED DURING EVALUATION

| Failure Category | Typical Observation |
|---|---|
| Etiology confusion | Arterial wounds classified as venous or vice versa |
| False infection detection | Infection predicted in clinically stable wounds |
| Over-testing | MRI or CT recommended without indication |
| Over-treatment | Unnecessary surgical debridement recommendations |
| Incorrect dressing selection | Dressing inconsistent with wound etiology |
| Procedural planning errors | Incorrect vascular or operative recommendations |

Among the open-source models, HuluMed showed the strongest qualitative performance. While it frequently identified major wound characteristics correctly, it struggled with advanced management decisions involving vascular intervention, debridement planning, and imaging recommendations.

The MedGemma models demonstrated several recurring failure patterns. These included excessive uncertainty, over-testing recommendations, inappropriate procedural suggestions, and difficulty distinguishing between arterial and venous etiologies. In several ischemic wound cases, MedGemma recommended debridement before restoration of adequate perfusion, representing clinically significant reasoning errors.

Overall, qualitative observations closely mirrored quantitative performance rankings. Models achieving higher accuracy scores also produced more clinically actionable and internally consistent responses, whereas lower-performing models exhibited greater uncertainty and management-planning errors.

## V. DISCUSSION

### A. Explaining the Performance Gap

The experimental data show a clear performance gap between generalist frontier systems and specialized medical models. The superior performance of ChatGPT (72.50%) and Claude (62.08%) indicates that massive multimodal pretraining and broad visual reasoning scale more effectively on specialized tasks than narrow, domain-specific tuning [7], [8], [19], [20]. Wound care requires identifying complex, ambiguous visual boundaries alongside varied clinical contexts [16], [17]. Frontier generalist systems use their extensive training scales to manage these varied contexts effectively, outperforming smaller open–source models that have only been exposed to specific, narrow medical datasets. A particularly notable finding is that neither MedGemma variant outperformed the non-medical Gemma 3 baseline. This suggests that domain-specific medical fine-tuning alone does not necessarily translate into improved wound-care reasoning, especially when tasks require integration of visual interpretation with complex management decisions.

### B. Understanding MedGemma's Performance Limitations

MedGemma's lower performance on this benchmark reveals key insights into model fine-tuning and alignment. The model's responses showed several consistent failure patterns:

- **Excessive Uncertainty and Hedging:** The model frequently used vague, over-cautious language that failed to provide actionable clinical decisions, resulting in zero points under the strict grading framework.
- **Linguistic Verbosity:** MedGemma regularly generated long explanations that lacked concise answers, leading to internal contradictions within its own text outputs.
- **Etiological Confusion:** The model often struggled to differentiate between complex vascular causes, confusing the distinct features of advanced arterial disease with chronic venous changes.

Overall, these findings indicate that current vision-language models derive substantially greater benefit from large-scale multimodal pretraining than from narrow domain-specific fine-tuning alone. While medical adaptation may improve terminology and clinical language generation, successful wound-care reasoning requires integration of visual interpretation, procedural knowledge, longitudinal decision making, and evidence-based treatment planning. The evaluated benchmark therefore highlights an important opportunity for future multimodal medical foundation models to jointly optimize visual understanding and clinical reasoning rather than focusing primarily on medical vocabulary or image recognition.

### C. Clinical Implications

Although frontier multimodal models substantially outperformed medically specialized open-source alternatives, none of the evaluated systems demonstrated sufficient reliability for autonomous clinical deployment. Even the highest-performing model, ChatGPT, achieved an overall accuracy of 72.50%, indicating that clinically significant errors remain common across complex wound-management tasks.

Interestingly, the largest performance differences emerged not in visual wound recognition but in downstream clinical reasoning. Most models were capable of identifying visible tissue characteristics such as necrosis, granulation tissue, and infection-related features. However, performance deteriorated considerably when translating these observations into management decisions involving vascular intervention, imaging selection, debridement strategy, or advanced wound therapies.

These findings suggest that current vision-language models should be viewed primarily as clinician-assistive technologies capable of supporting documentation, preliminary triage, and educational workflows rather than replacing specialist wound-care expertise. Human oversight therefore remains essential for safety-critical clinical decision making.

## VI. Limitations and Evaluation Boundaries

### A. Image-Only Evaluation Limitations

A primary limitation of this study is its reliance on static, image-only evaluation. Real-world clinical wound assessment requires an integrated review of the patient's broader medical context [17], which was unavailable in this benchmark:

- Complete longitudinal medical history, systemic status, and co-morbidities.
- Active laboratory values, including inflammatory markers, cultures, and glycemic controls.
- Objective non-invasive vascular assessments, such as ankle-brachial indices (ABI) and duplex ultrasounds.
- Physical examination findings like wound odor, palpation for deep bone exposure, and tissue texture.

Consequently, this study evaluates image-based clinical reasoning rather than definitive clinical diagnosis.

### B. Dataset and Reference Constraints

While the dataset was expanded to 20 highly varied clinical cases, the sample size remains a limitation for broader statistical validation. Additionally, the benchmark currently relies on a single consensus-driven reference standard. Future updates will incorporate multi-center inter-rater reliability checks to account for regional differences in wound-care practices.

## VII. Future Work

Future research will focus on expanding the evaluation framework across several areas:

- **Multicenter Datasets:** Validating performance on larger, ethnically diverse wound images to ensure robust visual evaluation across all skin tones.
- **Multimodal Integration:** Training models to process images alongside structured electronic health record data, laboratory values, and vascular flow metrics.
- **Human-in-the-Loop Workflows:** Testing these models as assistive decision-support tools to measure how VLM recommendations impact clinician triage speed and diagnostic accuracy.

## VIII. Conclusion

This study evaluated six vision-language models using a structured 20-case wound assessment benchmark covering 240 clinician-graded decisions. Frontier proprietary models outperformed specialized open–source alternatives, with ChatGPT achieving the highest overall accuracy (72.50%). Within this benchmark, large-scale generalist multimodal systems demonstrated stronger clinical reasoning performance than the evaluated medically specialized alternatives. The results further suggest that current model limitations arise less from visual recognition failures and more from difficulties translating visual findings into clinically appropriate management recommendations. While these systems show promise as assistive tools for documentation and clinical triage support, their current error rates mean they should function as interactive decision-support aids under direct clinician supervision rather than autonomous diagnostic platforms.